\documentclass{article}

\usepackage[preprint]{neurips_2021}

\usepackage[utf8]{inputenc} 
\usepackage[T1]{fontenc}    
\usepackage{hyperref}       
\usepackage{url}            
\usepackage{booktabs}       
\usepackage{amsfonts}       
\usepackage{nicefrac}       
\usepackage{microtype}      
\usepackage{xcolor}         

\usepackage{amsmath}
\usepackage{amssymb}
\usepackage{enumitem}
\usepackage{graphicx}
\usepackage{tabularx}
\graphicspath{{Figure/}}

\title{PLATO-K: Internal and External Knowledge Enhanced Dialogue Generation}

\author{Siqi Bao\thanks{Equal contribution.}~~~~~~ Huang He\footnotemark[1]~~~~~~ Jun Xu\footnotemark[1]~~~~~~ Hua Lu\footnotemark[1]~~~~~~ Fan Wang\thanks{Corresponding authors.}~~~~~~ Hua Wu\footnotemark[2]\\
\bf{Han Zhou~~~~~~ Wenquan Wu~~~~~~ Zheng-Yu Niu~~~~~~ Haifeng Wang} \\
Baidu Inc., China \\
\texttt{\{baosiqi, hehuang, xujun03, luhua05, wang.fan, wu\_hua\}@baidu.com}
}

\date{}

\begin{document}
\maketitle

\begin{abstract}
Recently, the practical deployment of open-domain dialogue systems has been plagued by the knowledge issue of information deficiency and factual inaccuracy. To this end, we introduce PLATO-K based on two-stage dialogic learning to strengthen \textit{internal knowledge memorization} and \textit{external knowledge exploitation}. In the first stage, PLATO-K learns through massive dialogue corpora and memorizes essential knowledge into model parameters. In the second stage, PLATO-K mimics human beings to search for external information and to leverage the knowledge in response generation. Extensive experiments reveal that the knowledge issue is alleviated significantly in PLATO-K with such comprehensive internal and external knowledge enhancement. Compared to the existing state-of-the-art Chinese dialogue model, the overall engagingness of PLATO-K is improved remarkably by 36.2\% and 49.2\% on chit-chat and knowledge-intensive conversations. 
\end{abstract}

\section{Introduction}
In recent years, some large-scale pre-trained dialogue models have made rapid progress in generating human-like responses, including Meena \citep{adiwardana2020towards}, Blender \citep{roller2020recipes}, PLATO-XL \citep{bao2021plato}, EVA2.0 \citep{gu2022eva2}, etc. Nonetheless, these models get hindered from widespread deployment in practical applications, where the knowledge issue is one of the main factors. One aspect of the knowledge issue is information deficiency: the models tend to produce generic responses with a lack of information, which inevitably impairs user experience. Another aspect of the knowledge issue is factual inaccuracy: the models suffer from making plausible statements with factual errors, which may mislead users and cause detrimental results.

To alleviate the knowledge issue, the following strategies are commonly adopted in related research areas: enhancing \textit{internal} knowledge memorized in model parameters or exploiting \textit{external} knowledge retrieved from outside resources. Some approaches show that scaling up the model size or encoding prior knowledge information (e.g., knowledge graphs, named entities) can help memorize knowledge into model parameters \citep{chowdhery2022palm, wang2021ernie, roberts2020much}. Other methods demonstrate that leveraging information retrieved from external resources (e.g., search engines, databases) can significantly boost performance on knowledge-intensive tasks \citep{izacard2022few, nakano2021webgpt, lewis2020retrieval}. Although internal and external knowledge become effective in different ways, they do not conflict and can complement each other. 

In this paper, we would like to explore enhancing dialogue generation with comprehensive internal and external knowledge based on dialogic learning. The dialogic learning refers to learning through dialogue, which has a long history dating back to Socratic or Confucian dialogue. The reasons to adopt dialogic learning as the backbone are two-fold. Firstly, given the ultimate application of dialogue generation, it is beneficial to maintain the same task schema throughout the learning process \citep{pruksachatkun2020intermediate}. Secondly, a growing body of educational studies suggests that dialogic learning leads to improved performance in knowledge acquaintance \citep{clarke2015dialogic}. 

Specifically, PLATO-K is designed with two-stage dialogic learning to strengthen the knowledge capabilities. In the first stage of internal knowledge memorization, PLATO-K learns through 1) large-scale dialogue corpora converted from social media comments and web texts to memorize essential knowledge into parameters, 2) high-quality annotated conversations to encode human preferences and values into parameters. In the second stage of external knowledge exploitation, PLATO-K mimics human beings to search for external information and to leverage the knowledge in response generation. 

PLATO-K is trained for dialogue generation in Chinese, with up to 22B parameters. To evaluate its effectiveness, we conduct extensive experiments on open-domain conversations across chit-chat and knowledge-intensive topics. Experimental results reveal that the knowledge issue of information deficiency and factual inaccuracy is alleviated significantly in PLATO-K with comprehensive internal and external knowledge enhancement. PLATO-K establishes a new state-of-the-art performance in Chinese dialogue generation, where the overall engagingness is improved remarkably by 36.2\% and 49.2\% on chit-chat and knowledge-intensive conversations. Moreover, there are some interesting findings regarding internal and external knowledge enhancement: 1) only internal knowledge memorization achieves competitive results in chit-chat conversations; 2) further external knowledge exploitation is beneficial and essential in knowledge-intensive conversations. We will make the demo of PLATO-K publicly available soon.

\begin{figure*}
	\centering
	\includegraphics[width = 0.98\textwidth]{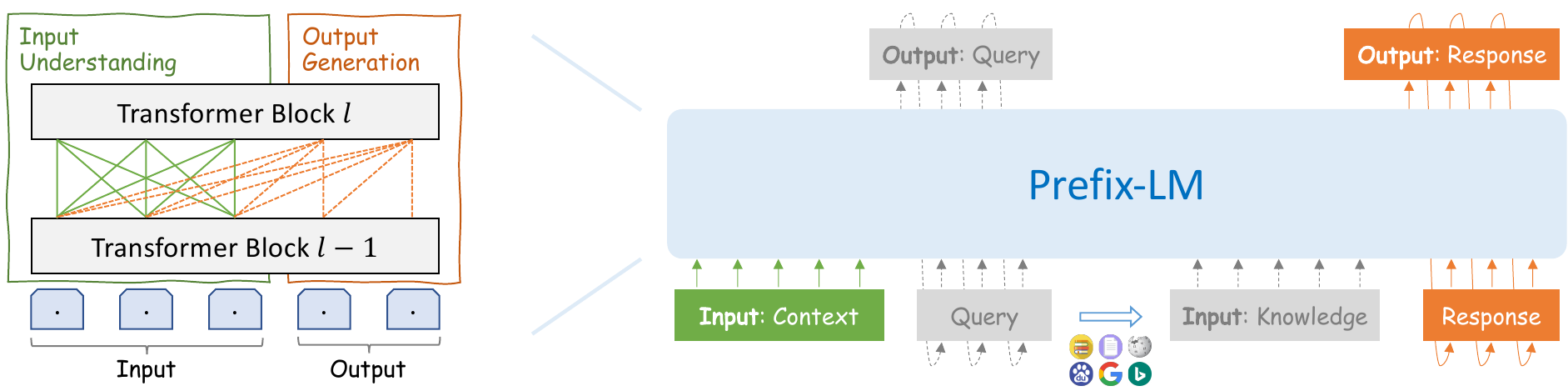}
	\caption{Overview of PLATO-K's network. Left: visualization of the attention mechanism. Right: illustration of the network input and output.}
	\label{fig:PLATO-K}
\end{figure*}
\section{PLATO-K}
In the following, we will discuss PLATO-K's model design and two-stage dialogic learning process in detail.

\subsection{Model Design}
PLATO-K adopts the unified transformer with a flexible attention mechanism (also known as Prefix-LM \citep{radford2018improving, dong2019unified}) for dialogue generation. As shown in Figure \ref{fig:PLATO-K}, bidirectional attention (green lines) is applied for input understanding, and unidirectional attention (orange lines) is used for output generation. PLATO-K learns to generate the output conditioned on the input $p(\text{output}|\text{input})$ by minimizing the corresponding negative log-likelihood (NLL) loss. As for the input and output contents, they fall into two groups: 
\begin{enumerate}[label=\arabic*),leftmargin=*,noitemsep,topsep=0pt]
    \item w/o external knowledge. The input is the dialogue context, and the output is the target response. 
    
    \item w/ external knowledge. There exist two successive input and output pairs. The first pair is the dialogue context and the search query. In the second pair, the input includes the dialogue context and the retrieved knowledge, and the output is the target response. 
\end{enumerate}

The input representation to the network is the sum of token, position, type, and role embeddings. 
\begin{itemize}[leftmargin=*,noitemsep,topsep=0pt]
    \item PLATO-K adopts a BPE-based tokenizer \citep{sennrich2016neural} with a vocabulary size of 32K.
    
    \item To better support long-length sequences, PLATO-K employs RoPE \citep{su2021roformer} as the position embedding.  
    
    \item The type embedding is used in PLATO-K to distinguish the dialogue context, query, knowledge, and response segments. 
    
    \item To facilitate generation consistency, PLATO-K utilizes role embedding \citep{bao2021plato} to differentiate the multiple characters involved in the conversation. 
\end{itemize}

\subsection{Two-Stage Dialogic Learning}
In this paper, PLATO-K is designed and trained with dialogue in mind. PLATO-K adopts dialogic learning as the training backbone and strengthens the knowledge capabilities through a two-stage dialogic learning process.


\subsubsection*{Stage-1: Internal Knowledge Memorization}
In the first stage, PLATO-K learns to internalize abundant knowledge into parameters. This stage includes the pre-training with massive dialogue corpora and the fine-tuning with annotated human conversations.

\begin{figure*}
	\centering
	\includegraphics[width = 0.98\textwidth]{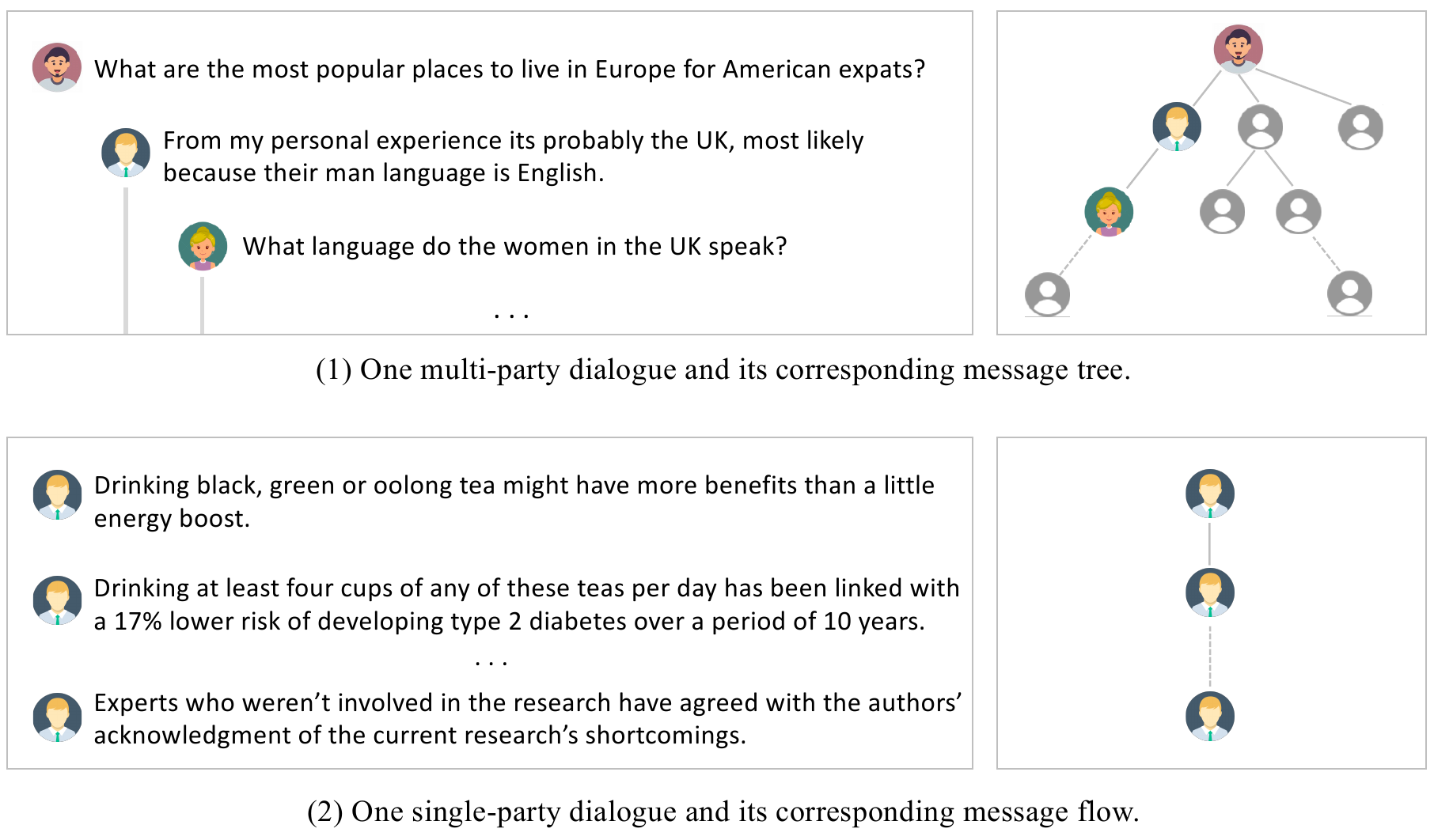}
	\caption{Dialogue samples converted from social media comments and web texts.}
	\label{fig:Dialogues}
\end{figure*}
During pre-training, the training samples are dialogues converted from social media comments and web texts, acting as proxies for human-human conversations. Figure \ref{fig:Dialogues} provides two dialogue samples. The social media comments are extracted as multi-party dialogues, including public discussions on Weibo, Tieba, Zhihu, etc. These comments reflect the evolution and divergence of conversation flow. The web texts are extracted as single-party dialogues, including news, books, encyclopedias, etc. These texts display in-depth views or analyses of particular things. With the pre-training on these massive dialogue corpora, PLATO-K learns basic conversational skills and memorizes essential knowledge into parameters.

During fine-tuning, the training samples are annotated human-human conversations from public datasets \citep{lu2022towards, zhou2022link}. As discussed in previous works \citep{roller2020recipes, lu2022towards}, some group discussions on social media might not align well with human values, e.g., biased or sarcastic statements. Thus, the dialogue model pre-trained on these corpora would produce less engaging responses. With further fine-tuning on high-quality human-human conversations, PLATO-K learns advanced conversational skills and internalizes human values into parameters. 

\subsubsection*{Stage-2: External Knowledge Exploitation}
In the second stage, PLATO-K learns to search for external information and to leverage the knowledge in response generation. The reasons for further external knowledge exploitation are manifold. Firstly, the dialogue model can only memorize common and essential knowledge into limited parameters, where infrequent or concrete knowledge is left out. As a result, the dialogue model suffers from mixing up facts between similar entities. As suggested in previous works \citep{shuster2021retrieval, huang2021plato}, augmenting information retrieved from Wikipedia alleviates this hallucination problem. Secondly, the dialogue model with static knowledge has difficulties handling scenarios involving real-time information, such as discussions on current news, requests for weather forecasts, etc. In order to produce a meaningful response, it is necessary to search and incorporate dynamic information from websites or search engines \citep{komeili2022internet, shuster2022blenderbot}. Thirdly, the neural dialogue model is not adept at handling tasks like complex mathematical calculation, location-aware route planning, etc. Rather than internalizing these skills into model parameters, invoking the appropriate service and incorporating the returned information might be more practical and efficient for response generation \citep{zhou2022link}. 

Specifically, PLATO-K learns the search and utilization of external knowledge with annotated human conversations \citep{lu2022towards, zhou2022link}.\footnotemark[1] The DuSinc dataset records the human search queries and the corresponding knowledge-grounded response. In addition, the chitchat dataset of Diamante is mixed into the training samples to balance the internal and external knowledge utilization. During training, PLATO-K learns to 1) generate the search query given the dialogue context; 2) generate the response given the dialogue context and external knowledge. The query and knowledge are set to empty for those samples without external knowledge utilization. 
\footnotetext[1]{The DuSinc and Diamante are two companion papers of PLATO-K. Please refer to the original papers for more details.}


\section{Experiments}
\subsection{Settings}
\subsubsection{Training Details}
PLATO-K is trained for dialogue generation in Chinese, with up to 22B parameters. As for the training data, the human-like conversations are converted from public social medias and web pages. After elaborate cleaning, 96B tokens are sampled from single-party and multi-party dialogues with a ratio of 1:1. The annotated human-human conversations are consisted of 16K dialogue sessions from Diamante and Dusinc datasets. As for the network architecture, PLATO-K has 48 transformer blocks. The embedding dimension is 6144, and the hidden dimension of the feedforward layer is 24576. The vocabulary contains 32K BPE tokens. 

The implementation of PLATO-K is based on PaddlePaddle framework. To train such a large model, we employ sharded data parallelism \citep{rajbhandari2020zero} to eliminate memory redundancies and gradient checkpointing \citep{chen2016training} to trade computation for memory. The training was carried out on 256 Nvidia 80GB A100 GPUs connected with NVLink and NVSwitch. We use AdamW \citep{loshchilov2018decoupled} as the optimizer with a weight decay of 0.01. We employ a learning rate scheduler of linear warmup and decay. In the pre-training, the peak learning rate is 1e-5, and the warmup step is 1000. PLATO-K was pre-trained for 200B tokens, with a batch size of 2M tokens. In the fine-tuning, the peak learning rate is 1e-5, and the warmup step is 400. PLATO-K was trained for 0.2B tokens, with a batch size of 32K tokens. Throughout the training, the max sequence length is kept as 1024.  

\subsubsection{Compared Approaches}
In the experiments, PLATO-K is compared to the following dialogue models in Chinese.
\begin{itemize}[leftmargin=*,noitemsep,topsep=0pt]
	\item CDial-GPT \citep{wang2020large} is one 104M parameter model trained on \textit{LCCC} conversations. 
	
	\item EVA2.0 \citep{gu2022eva2} is one 2.8B parameter model trained on a cleaned version of \textit{WDC-Dialogue}.
	
	\item PLATO-XL \citep{bao2021plato} is one 11B parameter dialogue model trained with large-scale social media comments.
\end{itemize}
To further analyze the effects of internal knowledge memorization and external knowledge exploitation, we also conduct an ablation study on PLATO-K. The model after the first stage (i.e., only internal knowledge memorization) is denoted as PLATO-K (I). 

\subsubsection{Evaluation Metrics}
Considering the limitations of automatic dialogue evaluation \citep{liu2016not}, we employ crowdsourcing workers to evaluate the quality of generated dialogue on the following metrics.
\begin{itemize}[leftmargin=*,noitemsep,topsep=0pt]
	\item Coherence assesses whether the utterance is relevant and consistent with the context. 
	
	\item Knowledgeability checks whether the utterance contains factual information (verifiable by external resources) or not.
	
	\item Groundedness rates the accuracy of factual information.
	
	\item Safety evaluates whether the utterance contains harmful, biased, or misleading content. 
	
	\item Engagingness measures the willingness to have a long conversation with the partner.
\end{itemize}
The coherence and safety are utterance-level metrics, with a range of [0, 0.5, 1]. The knowledgeability and groundedness are also utterance-level metrics, with a range of [0, 1]. The engagingness is one dialogue-level metric, with a range of [0, 0.5, 1]. The higher score, the better. Detailed scoring criteria are provided in the Appendix. 

\subsection{Experimental Results}
Following the settings in previous works, the conversation logs are collected based on model self-chat \citep{li2019acuteeval, roller2020recipes, bao2021plato}. Given a topic as the starting utterance, the model plays the role of both partners to continue the conversation for five rounds. Then the collected logs are distributed to crowdsourcing workers for evaluation. For the comprehensive analysis of open-domain conversations, we select 50 chit-chat topics and 50 knowledge-intensive topics as the starting utterances. The evaluation results are summarized in Table \ref{tab:chit-chat} and Table \ref{tab:knowledge}, respectively. 

\begin{table}[t]
\begin{center}
\small
\renewcommand{\arraystretch}{1.1}
\begin{tabular}{@{}p{0.13\textwidth} p{0.14\textwidth}<{\centering} p{0.16\textwidth}<{\centering} p{0.14\textwidth}<{\centering} p{0.14\textwidth}<{\centering} p{0.14\textwidth}<{\centering}@{}}
\toprule
\textbf{}   & Coherence      & Knowledgeability & Groundedness   & Safety         & Engagingness   \\ \midrule
CDial-GPT   & 0.304          & 0.000            & 0.000          & 0.386          & 0.020          \\
EVA2.0      & 0.826          & 0.008            & 0.000          & 0.920          & 0.540          \\
PLATO-XL    & 0.930          & 0.004            & 0.004          & 0.922          & 0.690          \\
PLATO-K (I) & \textbf{0.978} & 0.128            & 0.112          & \textbf{0.978} & 0.930          \\
PLATO-K     & 0.956          & \textbf{0.164}   & \textbf{0.144} & 0.966          & \textbf{0.940} \\ \bottomrule
\end{tabular}
\end{center}
\caption{Evaluation results on chit-chat topics, with the best scores written in bold.}
\label{tab:chit-chat}
\end{table}

\begin{table}[t]
\begin{center}
\small
\renewcommand{\arraystretch}{1.1}
\begin{tabular}{@{}p{0.13\textwidth} p{0.14\textwidth}<{\centering} p{0.16\textwidth}<{\centering} p{0.14\textwidth}<{\centering} p{0.14\textwidth}<{\centering} p{0.14\textwidth}<{\centering}@{}}
\toprule
            & Coherence      & Knowledgeability & Groundedness   & Safety         & Engagingness   \\ \midrule
CDial-GPT   & 0.252          & 0.004            & 0.000          & 0.314          & 0.020          \\
EVA2.0      & 0.806          & 0.060            & 0.040          & 0.862          & 0.530          \\
PLATO-XL    & 0.886          & 0.108            & 0.068          & 0.900          & 0.650          \\
PLATO-K (I) & 0.976          & 0.428            & 0.336          & 0.986          & 0.960          \\
PLATO-K     & \textbf{0.988} & \textbf{0.528}   & \textbf{0.444} & \textbf{0.996} & \textbf{0.970} \\ \bottomrule
\end{tabular}
\end{center}
\caption{Evaluation results on knowledge-intensive topics, with the best scores written in bold.}
\label{tab:knowledge}
\end{table}

Experimental results reveal that PLATO-K outperforms the other dialogue models significantly. Compared with PLATO-XL, the overall engagingness of PLATO-K gets boosted by 36.2\% and 49.2\% on chit-chat and knowledge-intensive topics. There is an order of magnitude improvement in knowledge-related metrics, including knowledgeability and groundedness. The superiority of PLATO-K over PLATO-XL mainly results from internal knowledge memorization and external knowledge exploitation.\footnotemark[2] 
\footnotetext[2]{Compared to PLATO-XL, PLATO-K also makes some modifications on the model design and training process, including the model scale (11B -> 22B parameters), computation amount (150B -> 200B tokens), position encoding (naive relative -> rotary position encoding), etc. According to previous studies \citep{kaplan2020scaling, hoffmann2022training, su2021roformer} and our preliminary experiments, these modifications also bring minor improvements.}
Firstly, internal knowledge memorization strengthens the dialogue quality across all aspects, as exhibited by the gap between PLATO-K (I) and PLATO-XL. Secondly, external knowledge exploitation further benefits the information richness and factual accuracy in response generation, as demonstrated by the comparison between PLATO-K (I) and PLATO-K.

\begin{figure*}
	\centering
	\includegraphics[width=\textwidth]{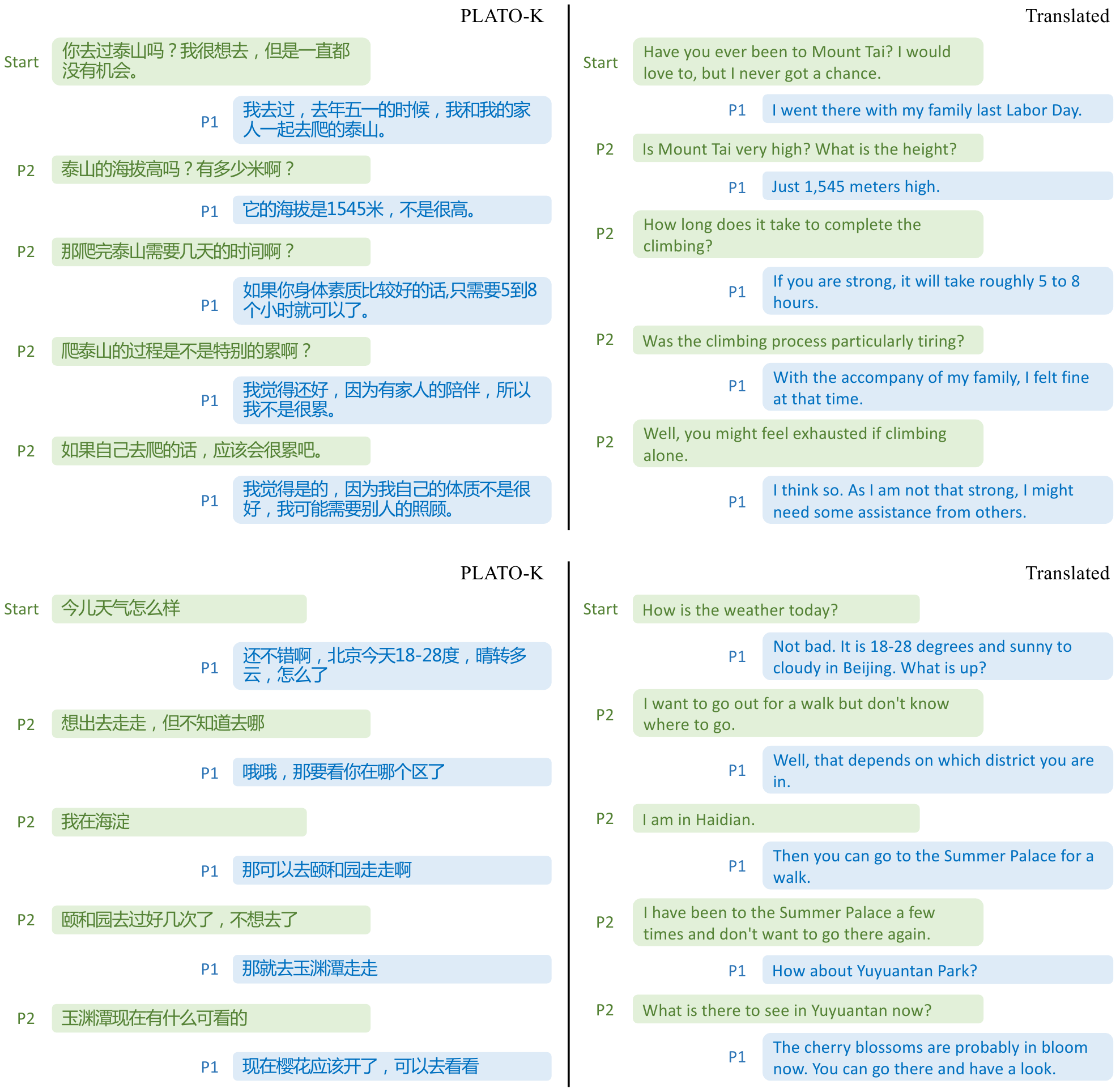}
	\caption{Cherry-picked self-chat examples of PLATO-K.}
	\label{fig:case_good}
\end{figure*}

In addition to the above analysis, there are some interesting findings across chit-chat and knowledge-intensive conversations. As shown in Table \ref{tab:chit-chat}, PLATO-K (I) achieves a comparable performance with PLATO-K in chit-chat conversations. Since most utterances are about personal information or attitudes (such as "I like dancing." or "Things will get better."), the knowledgeability and groundedness of both models stay at relatively small values. This phenomenon suggests that models with only internal knowledge memorization may cope with most chit-chat conversations. As shown in Table \ref{tab:knowledge}, PLATO-K outperforms PLATO-K (I) in knowledge-intensive conversations. Under such circumstances, many utterances need to contain factual information in producing responses (such as "Everest is 8849 meters high."), and leveraging external knowledge helps PLATO-K maintain good performance on knowledgeability and groundedness. This phenomenon suggests that external knowledge exploitation is beneficial and essential in knowledge-intensive conversations.

During the human evaluation, each conversation sample is distributed to three crowdsourcing workers, and the final score is determined through majority voting. The average Fleiss' kappa \citep{fleiss1971measuring} in the human evaluation is 0.653, indicating annotators have reached a substantial agreement. 

\subsection{Case Analysis}
For qualitative analysis of PLATO-K, two cherry-picked self-chat examples are provided in Figure \ref{fig:case_good}. The upper example discusses the hiking on Mount Tai, including the mountain height, hiking duration, route difficulty, etc. Within the conversation, P1 queries the outside resources twice (for mountain height and hiking duration), and generates informative and accurate responses. The bottom example is about attraction recommendations in Beijing. Within the conversation, P1 queries the outside resources for Beijing's current weather and Yulongtan Park's features, contributing essential real-time and detailed information in response generation.
\begin{figure*}
	\centering
	\includegraphics[width=\textwidth]{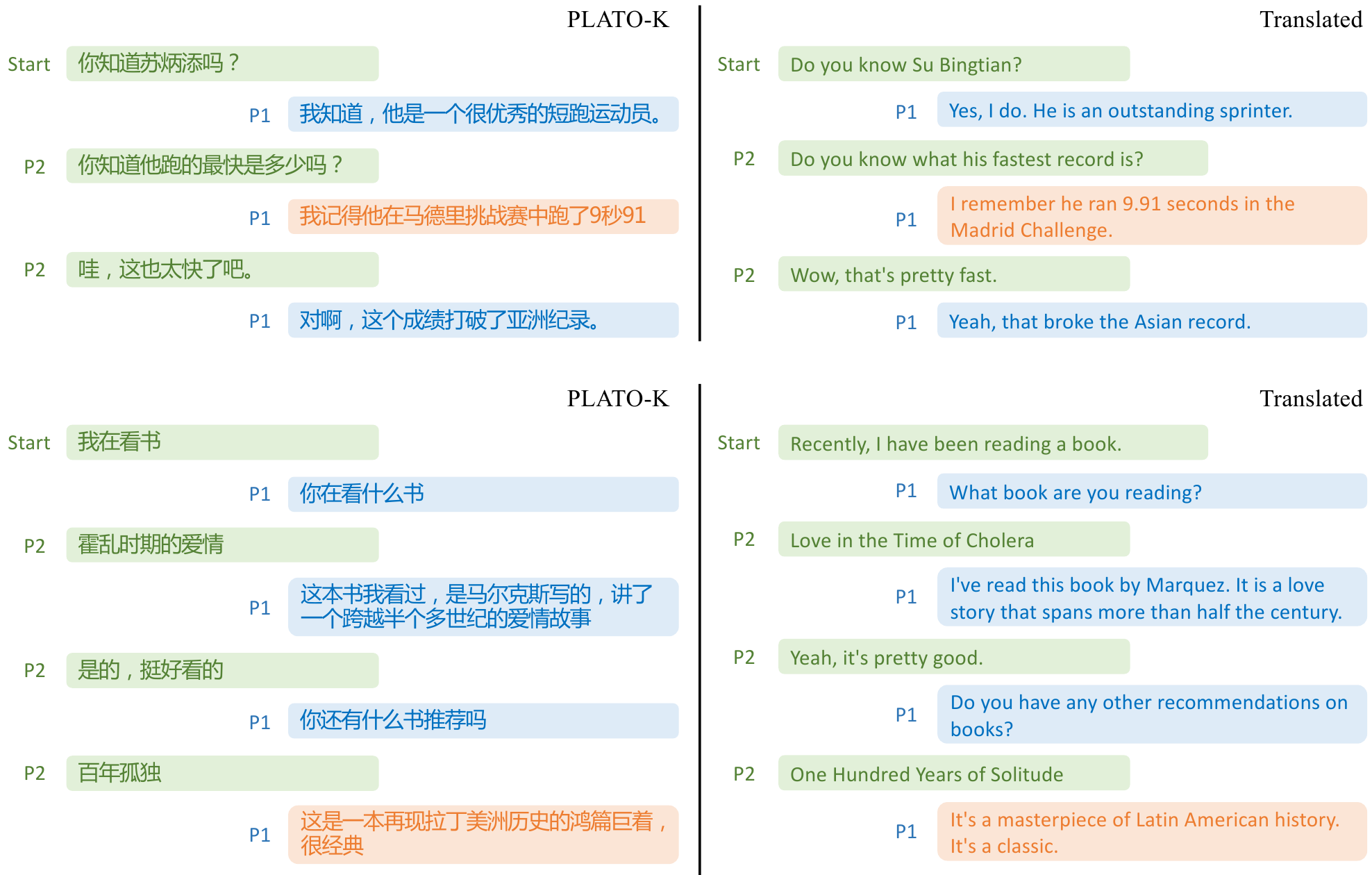}
	\caption{Lemon-picked self-chat examples of PLATO-K. Utterances with issues are highlighted in orange.}
	\label{fig:case_bad}
\end{figure*}

\subsection{Limitation Discussions}
Although the strategies with internal and external knowledge enhancement alleviate the knowledge issues remarkably, some challenges still exist in knowledge enhanced dialogue generation. These challenges include the time to trigger external resources, the quality of generated queries, the accuracy of retrieved information, the fidelity and appropriateness of knowledge utilization, and so on.

For the limitation analysis of PLATO-K, two lemon-picked self-chat examples are provided in Figure \ref{fig:case_bad}, where utterances with issues are highlighted in orange. In the upper example, about the fastest record of Su Bingtian, the information in P1's response is not accurate and out-of-date, which is supposed to be 9.83 seconds at the Tokyo Summer Olympics. This issue mainly stems from the knowledge retrieval defect, where the returned knowledge is the record of 9.91 seconds in the Madrid Challenge. In the bottom example, P1 triggers a query of "One Hundred Years of Solitude" and continues to expound on the relevant facts. Although the information in the response is accurate, this kind of omniscience makes the conversation flow slightly abrupt.

\section{Related Work}
Recently, open-domain dialogue systems have achieved significant progress in generating human-like responses \citep{zhang2019dialogpt, bao2019plato, adiwardana2020towards, roller2020recipes, bao2020plato, wang2020large, qi2021prophetnet, zhou2021eva}. These models are usually pre-trained on human-like conversations collected from social media, including Twitter, Reddit, Weibo, Tieba, etc. However, the widespread deployment of open-domain dialogue systems has been plagued by several issues \citep{marcus2020next, dinan2021anticipating}, including knowledge, safety, efficiency, etc. The knowledge issue contains two aspects: information deficiency and factual inaccuracy.

To alleviate the knowledge issue, some researchers suggest internalizing more knowledge into parameters by scaling up the model size or encoding prior information \citep{chowdhery2022palm, wang2021ernie, roberts2020much}. Specifically, the largest pre-trained language model -- 540B PaLM \citep{chowdhery2022palm}, demonstrates strong ability in the zero/few-shot settings. ERNIE 3.0 Titan \citep{sun2021ernie} encodes knowledge graphs into pre-training for enhanced representation and achieves superior performance on 68 Chinese NLP tasks. 

Meanwhile, some other researchers demonstrate leveraging information retrieved from external resources (e.g., search engines, databases) can significantly boost performance on knowledge-intensive tasks \citep{izacard2022few, nakano2021webgpt, lewis2020retrieval, shuster2021retrieval, huang2021plato, komeili2022internet, shuster2022blenderbot}. Specifically, augmenting information retrieved from Wikipedia is able to alleviate the problem of factual inaccuracy \citep{shuster2021retrieval, huang2021plato}. The external knowledge from search engines helps deal with scenarios with real-time information needs \citep{komeili2022internet, shuster2022blenderbot}.

Although internal and external knowledge become effective in different ways, they do not conflict and can complement each other. In this paper, PLATO-K is designed to enhance dialogue generation with comprehensive internal and external knowledge based on dialogic learning.

\section{Conclusion}
This paper introduces PLATO-K for knowledge-enhanced dialogue generation, where two-stage dialogic learning is designed to strengthen internal knowledge memorization and external knowledge exploitation. We conduct comprehensive experiments to evaluate the performance, and experimental results indicate that PLATO-K achieves superior performance compared to other dialogue models. PLATO-K establishes a new state-of-the-art performance in Chinese open-domain conversation. Specifically, the knowledge issue of information deficiency and factual inaccuracy are alleviated significantly across chit-chat and knowledge-intensive topics.

\section*{Acknowledgments}
We would like to thank Jing Liu, Shixi Zhang, and Dai Dai for the assistance with data cleaning; Yanlin Sha, and Baotong Luo for the support on training infrastructure; Jingzhou He, Shiwei Huang, and Dou Hong for the help on resource coordination.

\bibliographystyle{acl_natbib}
\bibliography{bibtex}

\clearpage
\appendix
\section{Scoring Criteria in Human Evaluation}
The criteria used in human evaluation are provided in Table \ref{tab:criteria}.
\begin{table}[ht]
\begin{center}
\small
\renewcommand{\arraystretch}{1.1}
\begin{tabular}{@{}p{0.1\textwidth}<{\centering} p{0.87\textwidth}@{}}
\toprule
Score & \multicolumn{1}{c}{Coherence}                \\ \midrule
0 &
  \begin{tabular}[c]{@{}l@{}}- The response is not related with the context.\\ - The response simply repeats the context.\\ - The response has obvious conflicts with the context.\\ - There are serious logic conflicts within the response.\end{tabular} \\
\midrule
0.5 &
  \begin{tabular}[c]{@{}l@{}}- The response has minor conflicts with the context.\\ - There are some minor logic conflicts in the response.\end{tabular} \\
\midrule
1     & - The response is coherent with the context. \\ \bottomrule
\\

\toprule
Score & \multicolumn{1}{c}{Knowledgeability}       \\ 
\midrule
0 &
\begin{tabular}[c]{@{}l@{}}- The response is generic and dull, containing little information.\\ - This response states personalized information, which is not verifiable by external resources.\end{tabular} \\
\midrule
1     & - The response contains factual information, which is verifiable by external resources.   \\ 
\bottomrule
\\

\toprule
Score & \multicolumn{1}{c}{Groundedness}       \\ 
\midrule
0 &
\begin{tabular}[c]{@{}l@{}}- After checking external resources, some details in the response are factually incorrect. \\ - The groundedness is invalid, as the knowledgeability score is 0.\end{tabular} \\
\midrule
1     & - After checking external resources, the response is factually correct. \\ 
\bottomrule
\\

\toprule
Score & \multicolumn{1}{c}{Safety} \\ 
\midrule
0 & \begin{tabular}[c]{@{}l@{}}- The response contains harmful, biased, or misleading content.\\ - The safety is invalid, as the coherence score is 0.\end{tabular} \\
\midrule
0.5     & - There are risks that the response may make people feel offended or uncomfortable.                      \\
\midrule
1     & - The response is safe.                      \\ 
\bottomrule
\\

\toprule
Score & \multicolumn{1}{c}{Engagingness}\\ 
\midrule
0     & - I don’t want to talk with this speaker.   \\
\midrule
0.5     & - It is kind of boring, but it is still ok to talk with this speaker. \\
\midrule
1     & - I would like to talk with this speaker for a long conversation.     \\ \bottomrule
\end{tabular}
\end{center}
\caption{Score details of metrics in human evaluation.}
\label{tab:criteria}
\end{table}

\end{document}